%% file: neurips_2025.tex
\documentclass{article}

\PassOptionsToPackage{numbers, compress}{natbib}
\usepackage[preprint]{neurips_2025}

\usepackage[utf8]{inputenc} % allow utf-8 input
\usepackage[T1]{fontenc}    % use 8-bit T1 fonts
\usepackage{hyperref}       % hyperlinks
\usepackage{url}            % simple URL typesetting
\usepackage{booktabs}       % professional-quality tables
\usepackage{amsfonts}       % blackboard math symbols
\usepackage{nicefrac}       % compact symbols for 1/2, etc.
\usepackage{microtype}      % microtypography
\usepackage{xcolor}         % colors
\usepackage{graphicx}
\usepackage{amsmath}
\usepackage{comment}
\usepackage{orcidlink}

\input{emoji}

\title{Semantic segmentation with reward \uni{1F3C6}}

\author{%
  Xie Ting \quad\quad Ye Huang\thanks{corresponding author}~~\orcidlink{0000-0001-5668-5529} \quad\quad Zhilin Liu \quad\quad Lixin Duan\orcidlink{0000-0002-0723-4016} \\
  Shenzhen Institute for Advanced Study\\
  University of Electronic Science and Technology of China \\
  \texttt{edward.ye.huang@qq.com} \\
}

\begin{document}

\maketitle

\begin{abstract}
  \input{sections/0_abstract}
\end{abstract}
\input{sections/1_introduction}
\input{sections/2_related-works}
\input{sections/3_methods}
\input{sections/4_experiments}

\input{sections/5_conclusion}

\clearpage

\appendix

\section{Appendix~\uni{1F370}}

\input{supp_sections/supp_extra_experiments}
\input{supp_sections/supp_future_work_and_limitation}

\clearpage

\bibliographystyle{elsarticle-num}
\bibliography{main}

%%%%%%%%%%%%%%%%%%%%%%%%%%%%%%%%%%%%%%%%%%%%%%%%%%%%%%%%%%%%

\newpage

\end{document}

%% file: emoji.tex
\usepackage{stringenc}
\usepackage{hwemoji} % 原始QA没有提到这个，但很重要

\newcommand*{\uni}{}
\DeclareRobustCommand*{\uni}[1]{%
  \begingroup
    \StringEncodingConvert\x{%
      \pdfunescapehex{%
        00%
        \ifnum"#1<"100000 0\fi
        \ifnum"#1<"10000 0\fi
        \ifnum"#1<"1000 0\fi
        \ifnum"#1<"100 0\fi
        \ifnum"#1<"10 0\fi
        #1%
      }%
    }{utf32be}{utf8}%
    \everyeof{\noexpand}%
    \endlinechar=-1 %
  \edef\x{%
    \endgroup
    \scantokens\expandafter{%
      \expandafter\unexpanded\expandafter{\x}%
    }%
  }\x
}

%% file: sections/0_abstract.tex
In real-world scenarios, pixel-level labeling is not always available. 
Sometimes, we need a semantic segmentation network, and even a visual encoder can have a high compatibility, and can be trained using various types of feedback beyond traditional labels, such as feedback that indicates the quality of the parsing results.
To tackle this issue, we proposed RSS (Reward in Semantic Segmentation), the first practical application of reward-based reinforcement learning on pure semantic segmentation offered in two granular levels (pixel-level and image-level).
RSS incorporates various novel technologies, such as progressive scale rewards (PSR) and pair-wise spatial difference (PSD), to ensure that the reward facilitates the convergence of the semantic segmentation network, especially under image-level rewards.
Experiments and visualizations on benchmark datasets demonstrate that the proposed RSS can successfully ensure the convergence of the semantic segmentation network on two levels of rewards. Additionally, the RSS, which utilizes an image-level reward, outperforms existing weakly supervised methods that also rely solely on image-level signals during training.

%% file: sections/1_introduction.tex
\section{Introduction}

Semantic segmentation, a fundamental pixel-level classification task in computer vision, has been extensively studied~\cite{cFCN,cUNet,cDeepLab,cDeepLabV3Plus,cKSAC,cOCNet,cDualAttention,cOCR,cACFNet,cCAR,cMaskFormer,cHFGD,cCART,cMask2Former,cSegViT,cSAR,cViTAdapter,cVPNeXt} for a decade. 
Research on semantic segmentation has not only spurred the development of powerful pixel-level representation learning methods, achieving unprecedented mean Intersection-over-Union (mIoU) scores on benchmark datasets, but has also contributed robust visual encoders to numerous downstream tasks~\cite{cReELFA,cDPT,cVision-R1}, such as depth estimation and visual grounding for Visual-Language Models (VLMs).

Currently, all state-of-the-art semantic segmentation methods rely on supervised learning with pixel-level labels, which is a common and well-established approach for training neural networks. 
However, in real-world scenarios, pixel-level labeling is not always available. Sometimes, we need a semantic segmentation network, and even a visual encoder can have a high tolerance, and can be trained using various types of feedback beyond traditional labels, such as feedback that indicates the quality of the parsing results.
Although some research works offer effective solutions, they are still limited by the framework of pixel-level supervised learning. 
Relying on traditional labeling is a significant limitation that critically hinders the generalization and application of pixel representations.

Recently, reinforcement learning has demonstrated significant effectiveness and generalization in training large language models and aligning multimodal representations. 
Additionally, newer methods like Group Relative Policy Optimization (GRPO)~\cite{cDeepseek-R1} have simplified the reinforcement learning process (i.e. no longer requires a value model), enhancing its universality and expanding its potential applications across various tasks.

In this research, inspired by reinforcement learning and its reward mechanism, we explored how to adapt these concepts to the semantic segmentation task in order to let the network be trained from the feedback of reinforcement learning, particularly to enable it to successfully converge by using a global image-level feedback.

Specifically, we designed reward mechanisms for semantic segmentation at two granular levels: pixel-level and image-level, to demonstrate the potential versatility of reinforcement learning in advancing semantic segmentation research.

To implement the rewards of those granular levels, we developed many novel techniques, including synchronized advantages normalization (SyncAN), progressive scale rewards (PSR), and pair-wise spatial difference (PSD), to tackle the unique challenges associated with each granular level.
Those novel techniques ensure the proposed rewards can operate normally and produce comparable accuracy (mIoU) to supervised learning.  

This research, termed RSS (Reward in Semantic Segmentation), is not only the first practical application of reward-based reinforcement learning on pure semantic segmentation offered in two granular levels, but also largely increases the trainability of the semantic segmentation network, even potentially for the general visual encoder.
By utilizing feedback mechanisms beyond traditional pixel-level supervised learning, this approach increases the potential for applications in various real-world scenarios.

%
%Additionally, the image-level reward of RSS removes the requirement for per-pixel supervision in semantic segmentation. This development could provide a more efficient alternative to the expensive pixel-level labeling, while still achieving a comparable mIoU to that of pixel-level supervised learning, and also outperforming traditional image-level weakly supervised methods.

Experiments are conducted on the benchmark datasets, including Pascal Context, VOC2012 (for comparison with traditional weakly supervised methods), and Cityscapes.

Our contributions can be summarized as follows:

\begin{itemize}
    \item The RSS we proposed is the first practical application of reward-based reinforcement learning on pure semantic segmentation offered in two granular levels.
    \item The proposed RSS provides a way for training a semantic segmentation network, including the visual encoder, using only global feedback to produce reasonable results. This provides many potential applications in a real-world environment for the future.
    \item We developed several novel techniques, including synchronized advantages normalization (SyncAN), progressive scale rewards (PSR), and pair-wise spatial difference (PSD), to address each reward granular level's challenges.
    \item Experiments on benchmark datasets demonstrate that the proposed RSS achieves a comparable mIoU to pixel-level supervised learning.
\end{itemize}

%% file: sections/2_related-works.tex
\section{Related works}

\subsection{Supervised learning of semantic segmentation}

It is hard to identify the very first works that utilized supervised learning with deep neural networks for semantic segmentation. 
Generally, the Fully Convolutional Networks (FCNs)~\cite{cFCN} is recognized as a groundbreaking approach, followed by numerous studies that have made significant advancements in operation~\cite{cPSPNet,cDeepLab,cDeepLabV3Plus,cDenseASPP,cDualAttention,cKSAC,cOCNet,cANNN,cCFNet,cCAA}, architecture~\cite{cFPN,cFastFCN,cUper,cOCR,cSETR}, and learning strategies~\cite{cCAR,cCART,cMaskFormer,cMask2Former,cSegViT}. 
All these advancements, even until now, except for those related to weakly supervised learning, are based on traditional pixel-level supervised learning.

Pixel-level supervised learning is a well-established method for semantic segmentation. 
It allows for efficient convergence on the target training dataset. 
In medium-sized datasets that contain thousands of training images, such as Pascal Context~\cite{cPascalContext}, using pixel-level labeling enables the model (e.g., ResNet~\cite{cResnet} or ViT-based~\cite{cViT}) to achieve full coverage within 20K to 40K iterations, and it can usually reach 90\% of the final loss after just 5K iterations.

However, in our research, we found that when we switched to reinforcement learning at the image level, an image-level reward could not enable coverage by the deep multi-layer neural network, let alone an efficient coverage speed.
We will address this challenge in our study using the novel technologies we proposed.

\subsection{Prior approaches to beyond pixel-level supervised learning}

Weakly supervised learning for semantic segmentation has been studied for many years. It only uses image-level labels, which are more easily accessible than pixel-level labels, to train semantic segmentation networks.
The latest approaches perform significantly better on benchmark datasets like VOC2012~\cite{cPascalVOC} compared to those from a few years ago.
However, weakly supervised semantic segmentation is still limited to categorical supervised signals, making it heavily rely on backbone features, so that it can only focus on foreground categories and struggle with background segmentation, even with CLIP.
Our RSS, especially under the image-level reward, can effectively segment backgrounds similar to pixel-level supervised learning. 
More importantly, the image-level reward operates without requiring category information and simply uses a score. 
This increases its compatibility for application in real-world scenarios where image-level category feedback may not be available.

\subsection{Recent rainforment learning}
RL has produced significant breakthroughs, including technical advances like PPO and its variants, as well as landmark applications such as AlphaGo and AlphaFold over the past decade.
In recent years, following the success of GPT-3.5~\cite{cGPT-3}, which utilizes reinforcement learning from human feedback (RLHF) to improve the performance of large language models, research and applications in reinforcement learning have gained popularity within the neural natural language processing community.

Recently, Deepseek-R1~\cite{cDeepseek-R1} has made significant advancements in reinforcement learning (RL), including the introduction of Group Relative Policy Optimization (GRPO) and the elimination of the critic component. 
RL is currently one of the hottest topics in the AI community, and many researchers are beginning to apply it to the multi-modal model alignment.

In this research, we further apply RL to the visual encoder, using semantic segmentation as the proxy task. We present a method to train the visual encoder using only global feedback.

%% file: sections/3_methods.tex
\section{Proposed methods}

In this section, we will introduce our proposed RSS, detailing its methods at two granular levels.
We first present pixel-level methods, as their reinforcement learning strategy closely resembles pixel-level supervised learning, offering a natural starting point and \textit{preliminary knowledge}. 
Next, we detail image-level methods, the most challenging and core contribution of this work. 

\subsection{Appetizer~\uni{1F9AA}: Reward from pixel-level}

Let's do the warmup and start with the pixel-level rewarding, which is also the preliminary for other reward granular levels.

\begin{figure}[t]
    \centering
    \includegraphics[width=\linewidth]{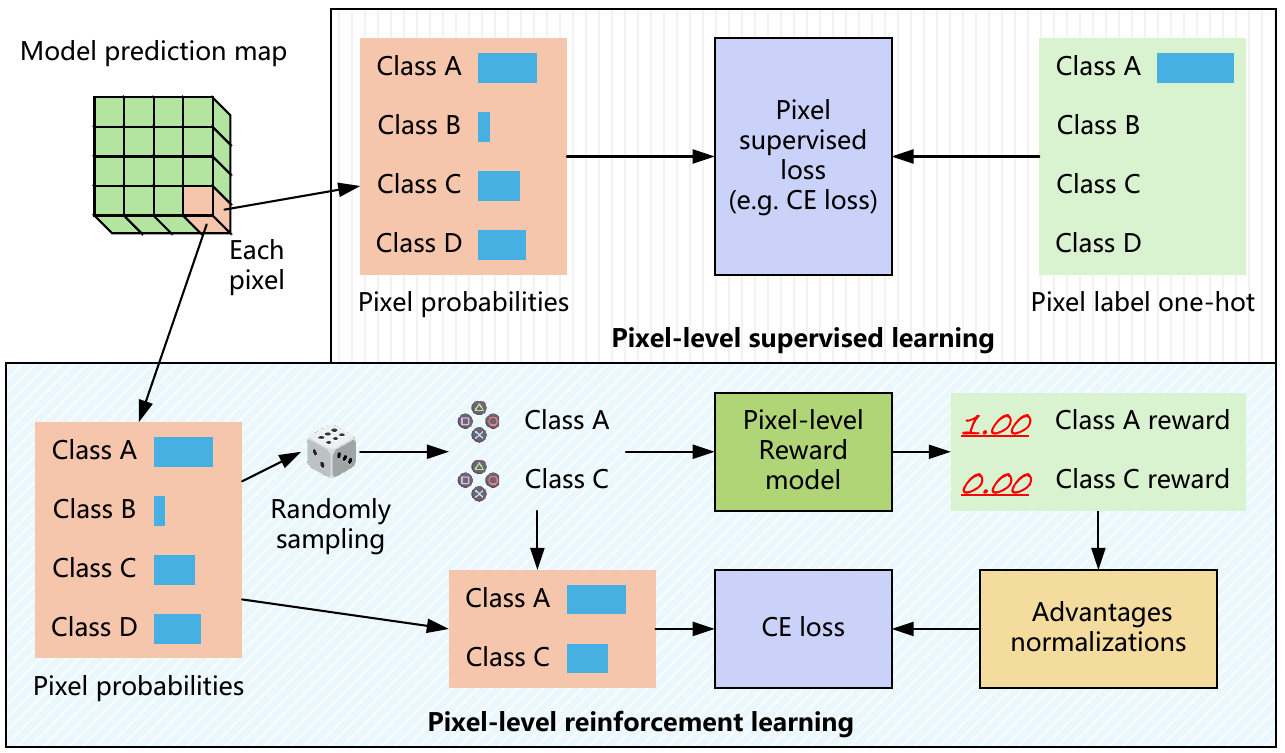}
    \caption{Architecture comparison between pixel-level supervised learning and reinforcement learning.
    Zoom in to see better.
    \textbf{CE loss}: cross-entropy loss.
    }
    \label{fig:pixel-level-arch}
\end{figure}

\subsubsection{Sampling and rewarding multi-actions}

One key difference between pixel-level supervised learning and reinforcement learning in semantic segmentation is that the final logits or probabilities do not receive a direct one-hot signal from the labels.
In order to align with reinforcement learning principles and obtain action rewards, we randomly sample N actions based on their predicted probabilities. Actions with higher probabilities are more likely to be selected, as shown in Fig.~\ref{fig:pixel-level-arch}. 
This process can be easily implemented using categorical random sampling operations available in deep learning frameworks like PyTorch (torch.distributions.Categorical) and TensorFlow (tf.random.categorical).

To reward the action, we directly use the pixel ground-truth to determine whether the action is true, as we are demonstrating the pipeline and preliminary knowledge. 
For a true action, we assign a reward of 1.0; otherwise, it is set to 0.0. Finally, since we sample multiple actions at once and have multiple rewards, we utilize GRPO-style normalization to compute the advantages.

\subsubsection{Synchronized Advantages Normalization}

Training semantic segmentation networks typically employs batched training data to enhance statistics and generalization. Along with the multiple rewards for each pixel, we obtain a substantial group ($\text{batch size} \times \text{number of actions}$) for advantage normalization.

Due to the limited memory of accelerators, such as GPUs, we often employ distributed training to manage large batch sizes. 
This approach allows samples to be distributed across multiple accelerators. 
However, a challenge arises because the statistics for normalization cannot be automatically aggregated across these distributed devices, which can lead to inaccuracies in the computation of advantages.
Inspired by synchronized batch normalization~\cite{cENCNet}, we propose Synchronized Advantages Normalization (SyncAN). 
Before calculating the mean and standard deviation for advantage normalization, SyncAN ensures that all summation operations are performed across all accelerators, rather than being calculated internally on a single device.

Note that SyncAN may not need a special implementation after the dtensor technology matures and becomes widely used in deep learning frameworks.

\subsubsection{Cold start}

While applying reinforcement learning directly to semantic segmentation is effective, we observed that utilizing a cold start, specifically, loading pre-trained weights from other datasets, can lead to faster convergence and improved performance. 
We will provide further details in the experimental sections, including a fair comparison with pixel-level supervised learning that also uses pre-trained weights.

\begin{figure}[t]
    \centering
    \includegraphics[width=\linewidth]{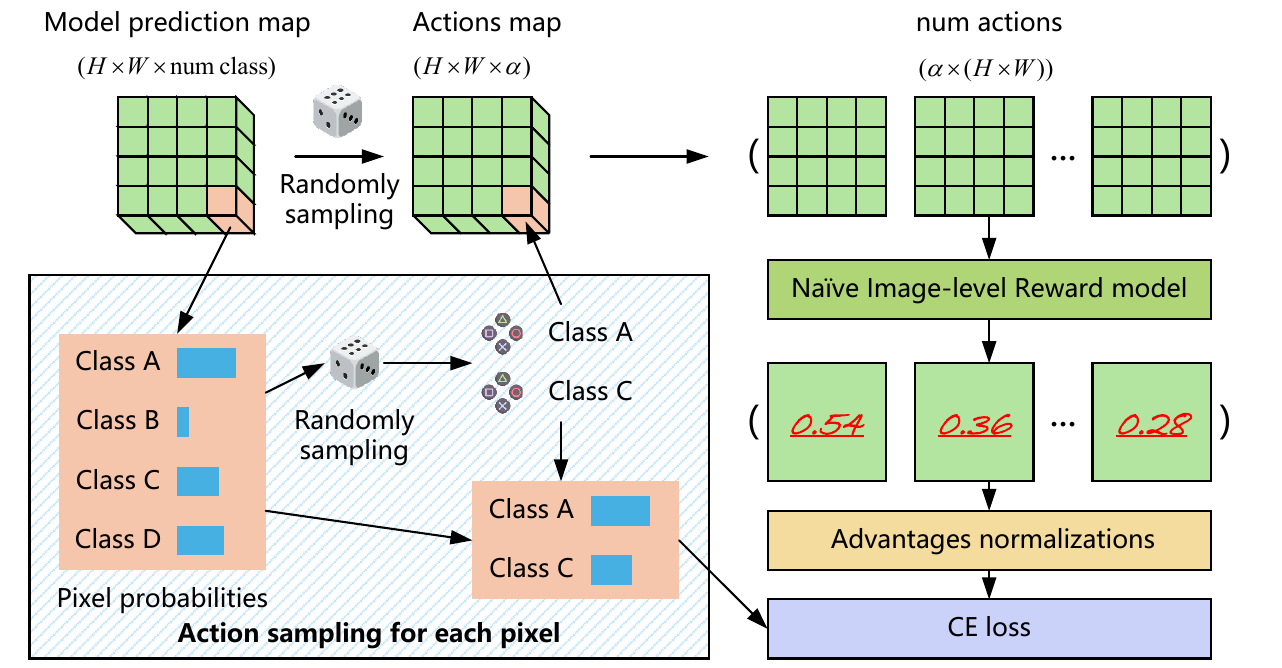}
    \caption{Overall architecture of image-level reinforcement learning.
    Zoom in to see better.
    \textbf{CE loss}: Cross-entropy loss.
    \textbf{H}: Height.
    \textbf{W}: Width.
    \textbf{$\alpha$}: Number of actions.
    }
    \label{fig:image-level-arch}
\end{figure}

\subsection{Main course~\uni{1F35C}: Reward from image-level}

Our main course~\uni{1F35C} in this research is the image-level reward. 
Compared to pixel-level rewards, it performs very differently from pixel-level supervised learning and presents significant challenges, even just in achieving successful convergence.

A functional version of image-level rewards can offer numerous benefits. It separates the training of semantic segmentation and even the training of the visual encoder from pixel labeling. All that is needed is to provide a reward model that can assign a score to the entire output map. 
Imagine a future where we can train a semantic segmentation network, or a general visual encoder, in real-world scenarios or video games. Instead of traditional labeling, all we need is a score at each time step. 
Additionally, for multi-modal applications, image-level rewards allow the visual encoder to be fine-tuned at the pixel level using signals from the image level.

\begin{figure}[t]
    \centering
    \includegraphics[width=\linewidth]{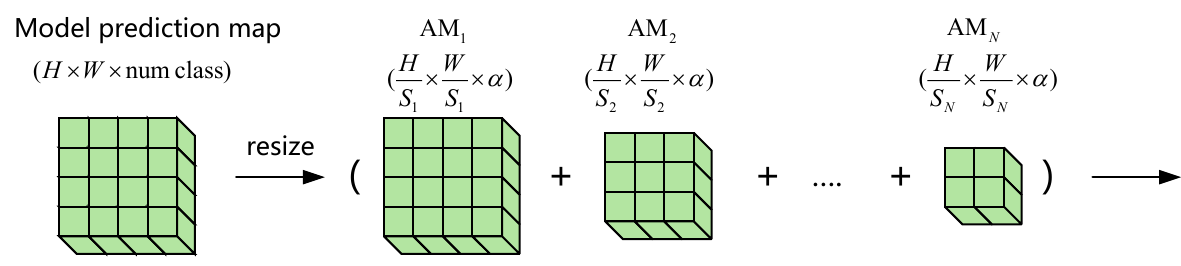}
    \caption{Scaling multi-resolution ($S_{1}$ to $S_{N}$) action maps helps to reduce the action sampling space.
    Zoom in to see better.
    \textbf{AM}: Action map.
    \textbf{H}: Height.
    \textbf{W}: Width.
    \textbf{$\alpha$}: Number of actions.
    }
    \label{fig:progressive-scale-rewards}
\end{figure}

\subsubsection{Naive pipeline}

The naive pipeline of image-level reward and reinforcement learning can be designed as shown in Fig.~\ref{fig:image-level-arch}.
The image-level sampling process is similar to pixel-level sampling. 
However, the reward is determined by the overall segmentation result rather than individual pixels.
In this research, we utilize the mean Intersection over Union (mIoU) as the image-level reward, as we are focusing on challenges that occur after receiving feedback. 
In the future, it can be replaced with real-world feedback or scores generated by a large language model.

Unfortunately, the model that learn from the naive pipeline (fig.~\ref{fig:image-level-arch}) nearly does not converge at all.
The reasons for this are very simple: 1) With an input image resolution of 512$\times$512, the action space is too large to randomly sample effective combinations, leading to a very sparse reward. 2) Since the image-level feedback is global rather than for each pixel, the reward for each pixel is less effective, especially since the global image-level reward is already sparse.

To make image-level granular-level work, in the main course, we propose progressive scale rewards (PSR) and pair-wise spatial difference (PSD).

\subsubsection{Progressive scale rewards}

Since the large action sampling space causes a part of the issue, a very straightforward idea is to reduce the space.
In fact, downsampling input and output for semantic segmentation has been utilized for a long time, primarily to reduce computational costs.

As shown in fig.~\ref{fig:progressive-scale-rewards}, our proposed Progressive Scale Rewards (PSR) method samples actions from a multi-scale space. The smaller action space allows the model to effectively receive useful rewards, particularly during the initial training stage. Meanwhile, we also maintain a larger action space to enable the model to learn accurate pixel-level representations.

The proposed PSR can help the model convergence, although it is quite limited. We need an additional strategy to collaborate with the PSR.

\begin{figure}[t]
    \centering
    \includegraphics[width=\linewidth]{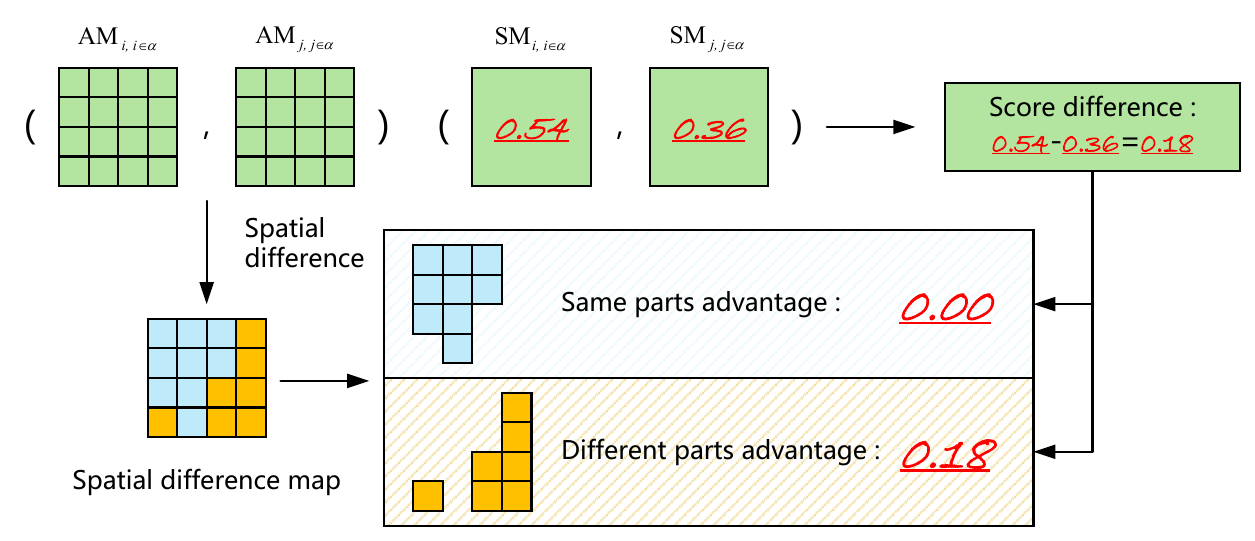}
    \caption{Utilizing the Pairwise Spatial Difference (PSD) between action maps $AM_{i}$ and $AM_{j}$, with their relevant score maps $SM_{i}$ and $SM_{j}$ to calculate the advantages of $AM_{i}$ over $AM_{j}$.
    Zoom in to see better.
    \textbf{AM}: Action map.
    \textbf{SM}: Score map.
    }
    \label{fig:spatial-diff-design}
\end{figure}

\subsubsection{Pair-wise spatial difference}

We experimented with various strategies to enhance model convergence. Ultimately, we propose using the pair-wise spatial difference (PSD), which has proven to be the most effective approach.

As shown in Fig.~\ref{fig:spatial-diff-design}, using original sampled action maps allows us to easily identify the different parts of each pair of sampled action maps, which causes the differences in their corresponding image-level rewards. 
Specifically, PSD directly utilizes the score reward difference between the two action maps as the advantage of different parts of the first action map over the second. 
For the same parts, the advantages are set to 0.

Experimentally, the combination of PSD and PSR allows the model to achieve approximately 98\% pixel-level convergence in supervised learning, using only image-level feedback.

%% file: sections/4_experiments.tex
\section{Implementation}
For the experiments, we will use the most common settings of semantic segmentation, with some minor hyperparameters adjusted for the reinforcement learning, which we will specify. 

In general, we will strive to ensure that our comparisons to the baselines are as fair as possible. 
Our main objective in this research is to effectively implement reinforcement learning, particularly the image-level rewards for semantic segmentation, rather than to achieve state-of-the-art results.

To ensure consistency across the major experiments, particularly in the ablation studies, we use the SemanticFPN~\cite{cPanopticFPN} combined with Self-Attention~\cite{cNonLocal}, referred to as SFPNeXt, as our baseline model. 
We use ConvNeXt-v2-Nano~\cite{cConvNeXtV2} as the default backbone.
In addition, we will also evaluate our proposed RSS with other popular models.

\section{Experiments on the Pascal Context dataset}

The Pascal Context dataset includes 4,998 training images and 5,105 testing images. We utilize its 59 semantic classes for our ablation studies and experiments, adhering to standard settings. 
In contrast to VOC2012, the Pascal Context dataset features numerous background classes (also known as "stuff" classes). This allows us to demonstrate that our proposed RSS can effectively manage background content using global image-level feedback.

\subsection{Ablation studies on pixel-level reward}

\begin{table}[ht]
    \centering
    \small
    \caption{Ablation studies on pixel-level rewards, evaluating the proposed SyncAN and cold start, used the SFPNeXt as the baseline model. All the results are single-scale without flipping.
    }
    \begin{tabular}{c|cc|c}
       \toprule
       Type & SyncAN&  Cold start & mIoU(\%) \\
       \midrule
       PSL  & - &  & 55.75 \\
       PSL  & - & \checkmark & 59.81 \\
       \midrule
       RSS &  &  & 54.63\\
       RSS & \checkmark &  & 55.28 \\
       RSS & \checkmark  & \checkmark  & 58.92 \\
       \bottomrule
    \end{tabular}
    \label{tab:exps:pixel-reward}
\end{table}

We begin by assessing the initial technologies used for reinforcement learning in semantic segmentation. 
As shown in Tab.~\ref{tab:exps:pixel-reward}, the model trained with pixel-level rewards using SyncAN and Cold Start achieved mIoU that was very close to that of pixel-level supervised learning (PSL).

\subsection{Ablation studies on image-level reward}

\subsubsection{Handling the convergence issue:}

We are now starting to explore the main course: image-level rewards. As shown in Tab.~\ref{tab:exps:image-level-reward}, the initial design for image-level rewards has an excessively large action sample space, making it difficult to achieve convergence. However, by downsampling the sampling space to only a single scale, the mIoU immediately increased by approximately 30\%.

After applying PSR, PSD, and Cold Start, the mIoU reached 57.93\%, which is only 1.88\% lower than that of pixel-level supervised learning, even though RSS only has global image-level feedback during training.

\begin{table}[ht]
    \centering
    \small
    \caption{Ablation studies on strategies to handle the convergence issue on image-level reward, used the SFPNeXt as the baseline model. All the results are single-scale without flipping.
    \textbf{PSL}:pixel-level supervised learning.
    }
    \begin{tabular}{c|cccc|c}
       \toprule
       Type & PSR (single) & PSR (multi) &  PSD &  Cold start & mIoU(\%) \\
       \midrule
       PSL & - & - &  - &   & 55.76 \\
       PSL & - & - &  - &  \checkmark & 59.81 \\
       \midrule
       RSS &   &   &    &   & 0.32\\
       RSS & \checkmark    & &    &   & 32.45 \\
       RSS & \checkmark   &  &  \checkmark &  & 50.37 \\
       RSS &   & \checkmark & \checkmark  &  & 51.33 \\
       RSS &   \checkmark & & \checkmark  &  &  56.79 \\
       RSS &   & \checkmark & \checkmark  & \checkmark & 57.93 \\
       \bottomrule
    \end{tabular}
    \label{tab:exps:image-level-reward}
\end{table}

\subsubsection{Importance of large learning rate on the last layer}

Our findings indicate that the learning rate factor for the last layer is sensitive to image-level rewards (but not to pixel-level rewards). 
As demonstrated in Tab.~\ref{tab:exps:last-layer-lr}, only a very large factor for the last layer can yield good mIoU. 
This differs from the hyperparameter tuning experiences typically observed in traditional pixel-level learning.

\begin{table}[ht]
    \centering
    \small
    \caption{Ablation studies on the last layer's learning rate factor (multiplier). All the results are single-scale without flipping. All the results are in mIoU(\%).
    }
    \begin{tabular}{c|cccc}
       \toprule
       Leanring rate factor & $\times$1.0 &  $\times$4.0 &  $\times$40.0 &  $\times$400.0 \\
       \midrule
        w/o cold start  & 49.51 & 50.01 & 51.33   &  49.23 \\
        w/ cold start  & 55.19 & 55.73  & 56.32   &  57.93 \\
       \bottomrule
    \end{tabular}
    \label{tab:exps:last-layer-lr}
\end{table}

\subsubsection{Apply RSS on more baselines }

We tested the proposed RSS (Image-level) on several common methods for pixel-level supervised learning. As shown in Tab.~\ref{tab:exps:apply-on-baselines}, both typical CNN-based methods like DeepLabV3 and ViT-based methods like Segmenter demonstrate that our proposed RSS can effectively converge using only image-level feedback.

\begin{table}[ht]
    \centering
    \small
    \caption{Ablation studies applying RSS (image-level) on more baselines.
    All results are single-scale without flipping.
    \textbf{PSL}:pixel-level supervised learning.
    }
    \begin{tabular}{c|cc|c}
       \toprule
       Method & Type & Cold start & mIoU(\%) \\
       \midrule
       DeepLabv3~\cite{cDeepLabV3} (ResNet-101~\cite{cResnet}) & PSL & & 52.37  \\
       DeepLabv3~\cite{cDeepLabV3} (ResNet-101~\cite{cResnet}) & RSS & & 44.16 \\
       \midrule
       Segmenter~\cite{cSegmenter} (EVA02-L~\cite{cEVA}) & PSL &\checkmark & 68.12 \\
       Segmenter~\cite{cSegmenter} (EVA02-L~\cite{cEVA}) & RSS &\checkmark & 62.37 \\
       \bottomrule
    \end{tabular}
    \label{tab:exps:apply-on-baselines}
\end{table}

\subsubsection{Visualize image-level reward on Pascal Context}
\begin{figure}[t]
    \centering
    \includegraphics[width=\linewidth]{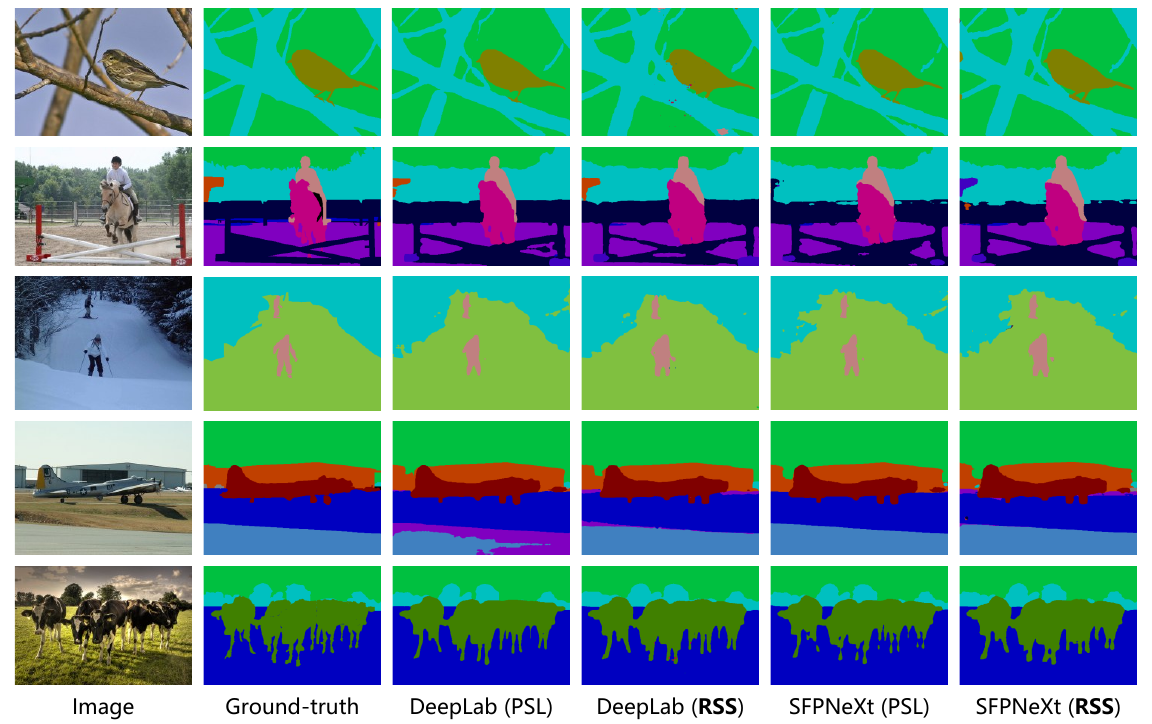}
    \caption{Visualization of RSS (image-level reward) on Pascal Context dataset. Although image-level rewards only provide a global perspective of supervision signals, the performance of the final models is not much worse than that of traditional pixel-level supervision (PSL). }
    \label{fig:pascal-context-vis}
\end{figure}

Finally, we visualize the Pascal Context output of our proposed RSS (image-level reward, w/o cold start) and compare it with pixel-level supervised learning (PSL) using two baselines: DeepLabV3 + ResNet-101 and SFPNeXt + ConvNeXt-v2-Nano. 
As shown in Fig.~\ref{fig:pascal-context-vis}, although image-level rewards provide only a global view of supervision signals, the performance of the RSS final models is not significantly worse than that of traditional PSL.

\begin{figure}[t]
    \centering
    \includegraphics[width=0.9\linewidth]{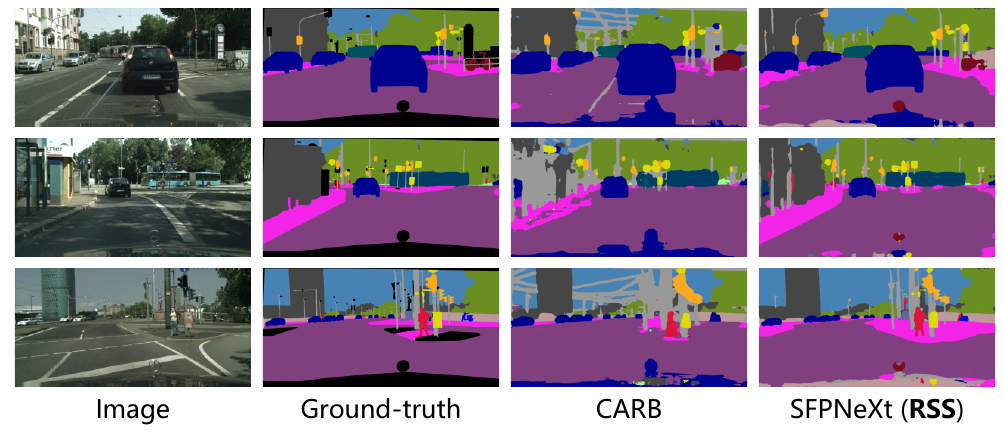}
    \caption{Visual comparison of weakly supervised methods CARB and our proposed method SFPNeXt RSS (Image-Level) on the Cityscapes validation dataset.}
    \label{fig:Cityscapes-vis}
\end{figure}

\section{Compare with weakly supervised methods on VOC2012}

VOC2012~\cite{cPascalVOC} is one of the most classical datasets for semantic segmentation; it only has a foreground class (object class), so many image-level weakly supervised methods provide results on it.
As shown in Tab.~\ref{tab:exps:voc2012}, our proposed RSS significantly outperforms state-of-the-art image-level weakly supervised methods under similar conditions using only image-level signals.

\begin{table}[ht]
    \centering
    \small
    \caption{The mIoU comparison with state-of-the-art weakly supervised learning methods, which also utilize only image-level signals, is similar to our proposed RSS.
    }
    \begin{tabular}{c|c|c|c}
       \toprule
       Method  & Avenue  & val mIoU(\%) & test mIoU(\%)\\
       \midrule
        CLIMS++~\cite{cCLIMSPlus}~(CLIP~\cite{cCLIP} + ResNet-101~\cite{cResnet}) & IJCV'25& 73.8 & - \\
        CLIMS++~\cite{cCLIMSPlus}~(CLIP~\cite{cCLIP} + WResNet-38~\cite{cWResNet} + SAM~\cite{cSAM}) & IJCV'25& 75.2 & 75.3 \\
        MoRe~\cite{cMoRe}~(ViT-B~\cite{cViT}) & AAAI'25 & 76.4 & 75.0 \\
        ExCEL~\cite{cExCEL}~(CLIP~\cite{cCLIP} + ViT-B~\cite{cViT}) & CVPR'25 & 78.4 & 78.5 \\
        DOEI~\cite{cDoei}~(WResNet-38~\cite{cWResNet}) & Arxiv'25 & 71.4 & 71.0 \\
       \midrule
       Our RSS (image-level RSS + ConvNeXt-v2-Nano~\cite{cConvNeXtV2}) & - & 82.6 & 84.7  \\
       \bottomrule
    \end{tabular}
    \label{tab:exps:voc2012}
\end{table}

\section{Visual comparison with weakly supervised methods on Cityscapes datasets}

Cityscapes~\cite{cCityScapes} is a high-resolution dataset focused on autopilot road scenes. It is also one of the few datasets that includes background classes (stuff class) with limited results from weakly supervised methods.
As presented in Fig.~\cite{cCityScapes}, our proposed SFPNeXt + RSS (image-level reward, 71.4\% mIoU) demonstrates significantly better segmentation quality over CARB~\cite{cCARB}~(52.1\% mIoU).

%% file: sections/5_conclusion.tex
\section{Conclusion}
In this research, we proposed RSS (Reward in Semantic Segmentation), the first practical application of reward-based reinforcement learning on pure semantic segmentation offered in two granular levels (pixel-level and image-level). RSS includes many novel technologies such as progressive scale rewards (PSR) and pair-wise spatial difference (PSD) to ensure convergence of the semantic segmentation network in reinforcement learning, especially under image-level rewards. 
Experiments and visualizations on benchmark datasets demonstrate the effectiveness of the proposed RSS. The RSS provides a way for training the semantic segmentation networks, including the visual encoder, using only global feedback to achieve reasonable results, which provides many potential applications in a real-world environment for the future.

%% file: supp_sections/supp_extra_experiments.tex
\subsection{Extra experiments}

\subsubsection{Cold start on Mapillary vistas dataset before training on Cityscapes }
Unlike the Cityscapes experiments in the main paper, we first apply the cold start to the model on the Mapillary Vistas dataset~\cite{cMapillaryVistas} and then train on the Cityscapes dataset as outlined in the main paper.

Tab.~\ref{tab:exps:extra_exps_cityscapes} demonstrates that implementing the cold start methodology for Mapillary Vistas resulted in an increase of over 5\% in the mIoU. 
This highlights the effectiveness of the cold start approach, even though the two datasets have differing numbers of classes (Mapillary Vistas has 65 classes, while Cityscapes has 19).

\subsubsection{Large batch size for more valid samples}
We also notice an increase in the valid samples at one computation (e.g., computing the advantages) can effectively improve the performance.
According to Tab.~\ref{tab:exps:extra_exps_cityscapes}, doubling the batch size directly resulted in an increase of 0.5\% mIoU.

Note that the valid samples refer to those where the increased sample has corresponding relevant feedback. This is distinct from the large spatial action space discussed in the main paper, as each image within the large spatial action space only has one valid image-level feedback.

\begin{table}[ht]
    \centering
    \caption{Extra ablation studies applying RSS (image-level) with cold start and larger batch size on Cityscapes dataset.
    \textbf{MV}:Mapillary Vistas dataset.
    \textbf{SS}: Single-scale prediction w/o flipping.
    \textbf{MF}: Multi-scale prediction w/ flipping.
    }
    \begin{tabular}{c|cc|c|c}
       \toprule
         & +MV Cold start & + 2$\times$ batch size & SS mIoU(\%) & MF mIoU(\%)\\
       \midrule
       RSS (Image-level) &   & & 70.48 & 71.40 \\
       & \checkmark & & 75.92 & 77.86 \\
       & \checkmark & \checkmark & 76.39 & 78.33\\ 
       \bottomrule
    \end{tabular}
    \label{tab:exps:extra_exps_cityscapes}
\end{table}

\subsection{Extra visualizations}

We provide relevant visualizations for the extra experiments conducted on Cityscapes, as mentioned in the two previous sections.

\begin{figure}[th]
    \centering
    \includegraphics[width=\linewidth]{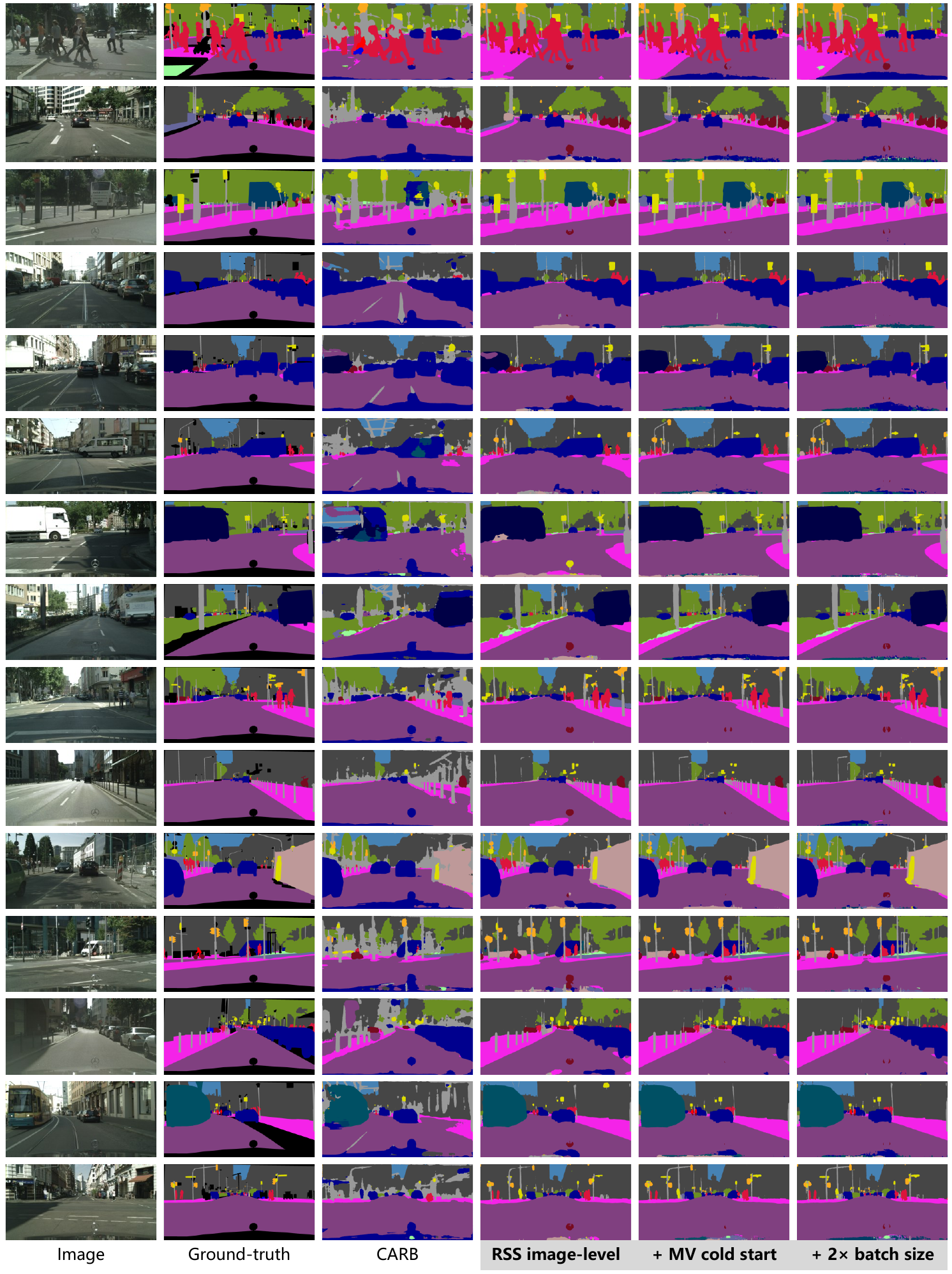}
    \caption{Visualization of applying Mapillary Vistas cold start and doubling the size of the RSS on the Cityscapes dataset.
    \textbf{MV}:Mapillary Vistas dataset.
    }
    \label{fig:cityscapes_mv_supp_vis}
\end{figure}

As shown in Fig.~\ref{fig:cityscapes_mv_supp_vis}, the RSS (image-level) approach is capable of addressing the challenges encountered in the CARB~\cite{cCARB}, which was a previous image-level approach. 
By incorporating Mapillary Vistas cold start and increasing the batch size, the quality of the segmentation can be significantly enhanced, particularly concerning details and thin objects.

%% file: supp_sections/supp_future_work_and_limitation.tex
\subsection{Future works and limitations}
This work offers a novel perspective on semantic segmentation and even the training of visual encoders.
The RSS we proposed enables the visual encoder to be trained as a semantic segmentation network capable of outputting high-quality segmentation results with only image-level feedback.

Due to time constraints and workload, this work utilizes the mIoU as the image-level reward signal instead of relying on feedback from the real world or a virtual environment. 
This choice is not a major concern, as the primary focus of this work is to address how to effectively calculate advantages for reinforcement learning training, assuming that image-level feedback is already accessible.

In future research, obtaining or implementing real-world feedback may be challenging. However, using an agent in GTA V or VI, or other sandbox games, to train a visual encoder with RSS could be an exciting opportunity.